\documentclass[letterpaper]{article} 
\usepackage{aaai24}  
\usepackage{times}  
\usepackage{helvet}  
\usepackage{courier}  
\usepackage{graphicx} 
\usepackage[hyphens]{url}  
\usepackage{natbib}  
\usepackage{caption} 
\usepackage{bm}
\usepackage{algorithm}
\usepackage{algorithmic}
\usepackage{amsmath}
\usepackage{booktabs}
\usepackage{multirow}
\usepackage{arydshln}
\usepackage{newfloat}
\usepackage{listings}
\DeclareCaptionStyle{ruled}{labelfont=normalfont,labelsep=colon,strut=off} 
\floatstyle{ruled}
\newfloat{listing}{tb}{lst}{}
\floatname{listing}{Listing}
\title{Learning Dense Correspondence for NeRF-Based Face Reenactment}
\author{
    Songlin Yang\textsuperscript{\rm 1,}\textsuperscript{\rm 2}, Wei Wang\textsuperscript{\rm 2}\thanks{Corresponding author.}, Yushi Lan\textsuperscript{\rm 3}, Xiangyu Fan\textsuperscript{\rm 4}, Bo Peng\textsuperscript{\rm 2}, Lei Yang\textsuperscript{\rm 4}, Jing Dong\textsuperscript{\rm 2}
}
\affiliations{
    \textsuperscript{\rm 1}School of Artificial Intelligence, University of Chinese Academy of Sciences, China\\
    \textsuperscript{\rm 2}CRIPAC \& MAIS, Institute of Automation, Chinese Academy of Sciences, China\\
    \textsuperscript{\rm 3}S-Lab, Nanyang Technological University, Singapore\\
    \textsuperscript{\rm 4}SenseTime, China\\

    yangsonglin2021@ia.ac.cn, \{wwang, bo.peng, jdong\}@nlpr.ia.ac.cn,\\ yushi001@e.ntu.edu.sg, \{fanxiangyu, yanglei\}@sensetime.com
}

\usepackage{bibentry}

\begin{document}

\maketitle

\begin{abstract}
Face reenactment is challenging due to the need to establish dense correspondence between various face representations for motion transfer. Recent studies have utilized Neural Radiance Field (NeRF) as fundamental representation, which further enhanced the performance of multi-view face reenactment in photo-realism and 3D consistency. However, establishing dense correspondence between different face NeRFs is non-trivial, because implicit representations lack ground-truth correspondence annotations like mesh-based 3D parametric models (e.g., 3DMM) with index-aligned vertexes. Although aligning 3DMM space with NeRF-based face representations can realize motion control, it is sub-optimal for their limited face-only modeling and low identity fidelity. Therefore, we are inspired to ask: \textbf{\textit{Can we learn the dense correspondence between different NeRF-based face representations without a 3D parametric model prior?}} To address this challenge, we propose a novel framework, which adopts tri-planes as fundamental NeRF representation and decomposes face tri-planes into three components: canonical tri-planes, identity deformations, and motion. In terms of motion control, our key contribution is proposing a Plane Dictionary (\textbf{PlaneDict}) module, which efficiently maps the motion conditions to a linear weighted addition of learnable orthogonal plane bases. To the best of our knowledge, our framework is the first method that achieves one-shot multi-view face reenactment without a 3D parametric model prior. Extensive experiments demonstrate that we produce better results in fine-grained motion control and identity preservation than previous methods.
\end{abstract}

\begin{figure*}
    \centering
    \includegraphics[scale=0.53]{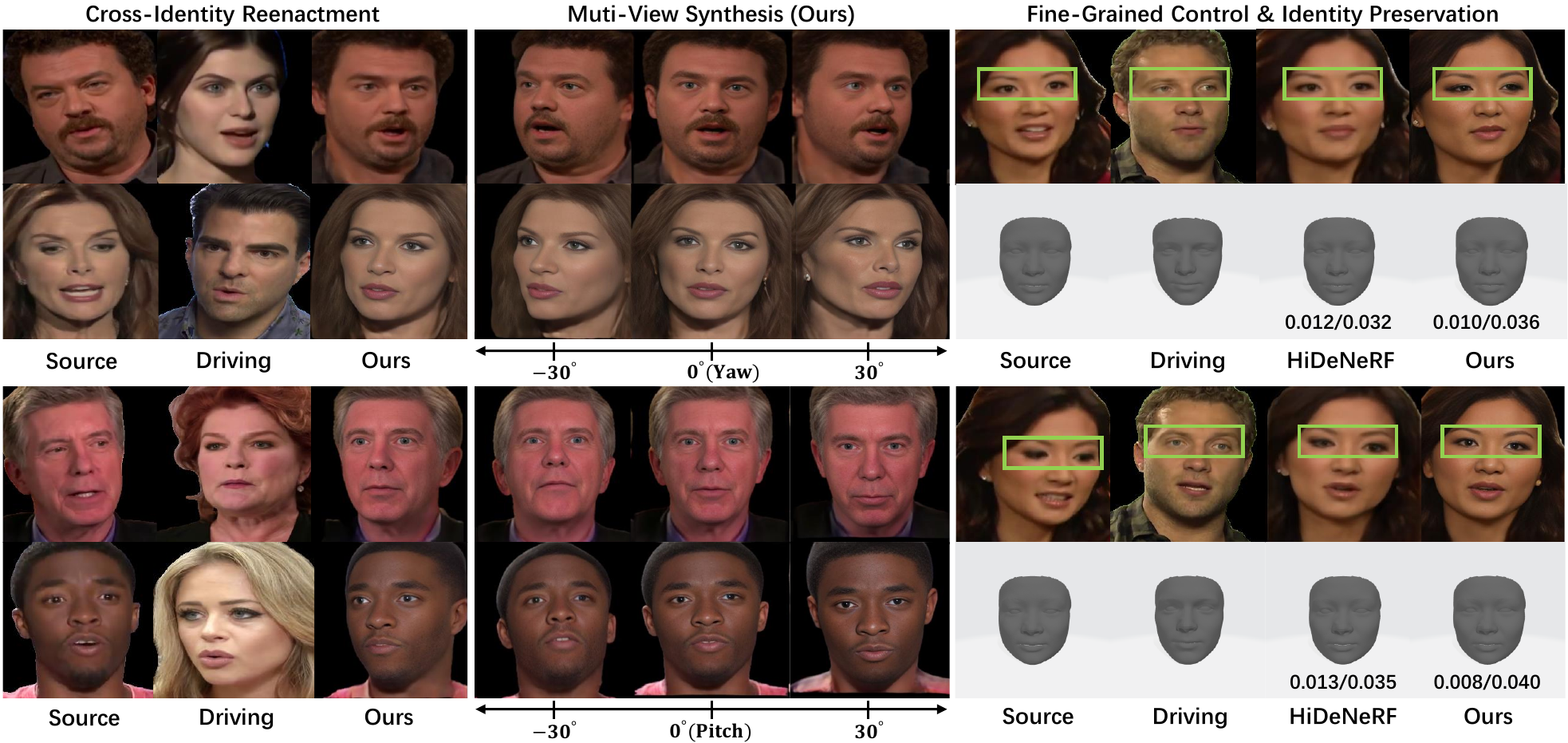}
    \caption{Results of one-shot multi-view face reenactment. We present both cross-identity reenactment and multi-view synthesis at various yaw and pitch angles. Through comparisons with state-of-the-art HiDeNeRF, we illustrate that our PlaneDict module excels in fine-grained motion control, particularly for non-facial elements such as eyes, and offers better identity preservation (vertex distance from source identity$\downarrow$/ vertex distance from driving identity$\uparrow$) than utilizing 3DMM as correspondence.}
    \label{introduction}
    
\end{figure*}

\section{Introduction}

One-shot face reenactment~\cite{hong2022depth} aims to utilize motion conditions from the driving image, such as facial expressions and head poses, to animate the face in the source image. The main challenge is establishing dense correspondence between different face representations to transfer motion conditions. Recent studies~\cite{Li_2023_CVPR,ma2023otavatar} have utilized Neural Radiance Field (NeRF)~\cite{mildenhall2021nerf} as fundamental representation, which further enhanced the performance of multi-view face reenactment in photo-realism and 3D consistency. 

However, establishing dense correspondence between different face NeRFs is non-trivial. Unlike mesh-based representations which have index-aligned vertexes as ground-truth correspondence annotations, the NeRF-based representations lack an explicit surface descriptor that constructs correspondence of spatial points~\cite{lan2022correspondence}. Although introducing 3D parametric models (e.g., 3DMM~\cite{blanz1999morphable}, FLAME~\cite{li2017learning}, and DECA~\cite{feng2021learning}) as motion conditions make it feasible to achieve explicit motion control for cross-identity face reenactment~\cite{zeng2022fnevr,ma2023otavatar,Li_2023_CVPR}, aligning mesh-based parametric space with latent space of NeRF-based generative models brings a significant optimization burden. Additionally, the 3D parametric models themselves have some limitations, such as their focus being predominantly on the facial region, requiring additional processing for hair and eyes. These limitations inspire us to ask: Can we learn the dense correspondence between different NeRF-based face representations without a 3D parametric model prior? 

To address the challenge of learning dense correspondences between NeRF-based face representations, the first issue is the selection of NeRF representations. The vanilla NeRF~\cite{mildenhall2021nerf} employs an MLP network to capture the spatial information of the target object, which tends to suffer from overfitting and can lead to a loss of 3D consistency when animating the representation network. In order to strike a balance between 3D consistency and animation capabilities, we have opted to utilize the tri-plane representation proposed by EG3D~\cite{chan2022efficient} as our fundamental NeRF representation, which adopts three spatially-orthogonal plane feature maps to represent an object. This choice allows us to maintain 3D consistency within the tri-plane representation itself, while also leveraging the strong modeling capacity of the StyleGAN-based~\cite{karras2020analyzing} generator to handle feature deformations.

In this work, we propose a novel framework that can realize one-shot multi-view face reenactment, as shown in Fig.~\ref{introduction}. Our framework utilizes tri-planes as fundamental NeRF representation and decomposes face tri-planes into three components: canonical tri-planes, identity deformations, and motion. The plane feature deformations regarding motion conditions differ from those caused by identity conditions since the rules governing motion are shared between various identities. Thus, we design a Plane Dictionary module, referred to as \textbf{PlaneDict}, to efficiently maps motion conditions to a linear weighted addition of learnable orthogonal plane bases. Extensive experiments demonstrate that our method achieves better results in fine-grained motion control and identity preservation than previous work.

\textbf{To summarize, the contributions of our approach are:}
\begin{itemize}
    \item We propose a novel decomposition method of face tri-plane representation, making it suitable for learning the dense correspondence between different face tri-planes and realizing motion transfer.
    \item We propose a Plane Dictionary (\textbf{PlaneDict}) module for tri-plane representation, which efficiently maps motion conditions to a linear weighted addition of learnable orthogonal plane bases.
    \item To the best of our knowledge, we propose the first method to achieve one-shot multi-view face reenactment without a 3DMM prior, which achieves better results in fine-grained motion control and identity preservation than previous work. 
\end{itemize}

\section{Related Work}

\subsection{Face Implicit Representation}

Compared with 2D~\cite{liu2015deep} and 3D parametric representation~\cite{li2017learning}, 3D implicit representation has advantages of photo-realism and 3D-consistency. Previous work based on 3D scene representation has tried to use Neural Radiance Field~\cite{gu2021stylenerf,yang2023context}, Signed Distance Field~\cite{or2022stylesdf,ma2023semantic}, and Tri-Planes~\cite{chan2022efficient} to model face as static objects. Considering the dynamic synthesis requirements of faces, two strategies have been proposed: First, constructing the deformable neural radiance fields, such as NeRFies~\cite{park2021nerfies} and HyperNeRF~\cite{park2021hypernerf}, which maps each observed point into a canonical space through a continuous deformation field, but it tends to handle small movements or person-specific rendering. Second, adopting NeRF with 3DMM~\cite{li2017learning} prior for explicit motion control, such as RigNeRF~\cite{athar2022rignerf}, NeRFace~\cite{gafni2021dynamic}, MoFaNeRF~\cite{zhuang2022mofanerf}, OmniAvatar~\cite{xu2023omniavatar}, and some 3D GAN Inversion methods~\cite{lin20223d,lan2023self,yang2023designing}. However, the dense correspondence provided by 3DMM has limitations in non-facial regions (e.g., eyes and hair) and brought an optimization burden to align the 3DMM representation and NeRF-based latent space. Therefore, a better dense correspondence of different 3D implicit representations is needed.

\begin{figure*}
    \centering
    \includegraphics[scale=0.54]{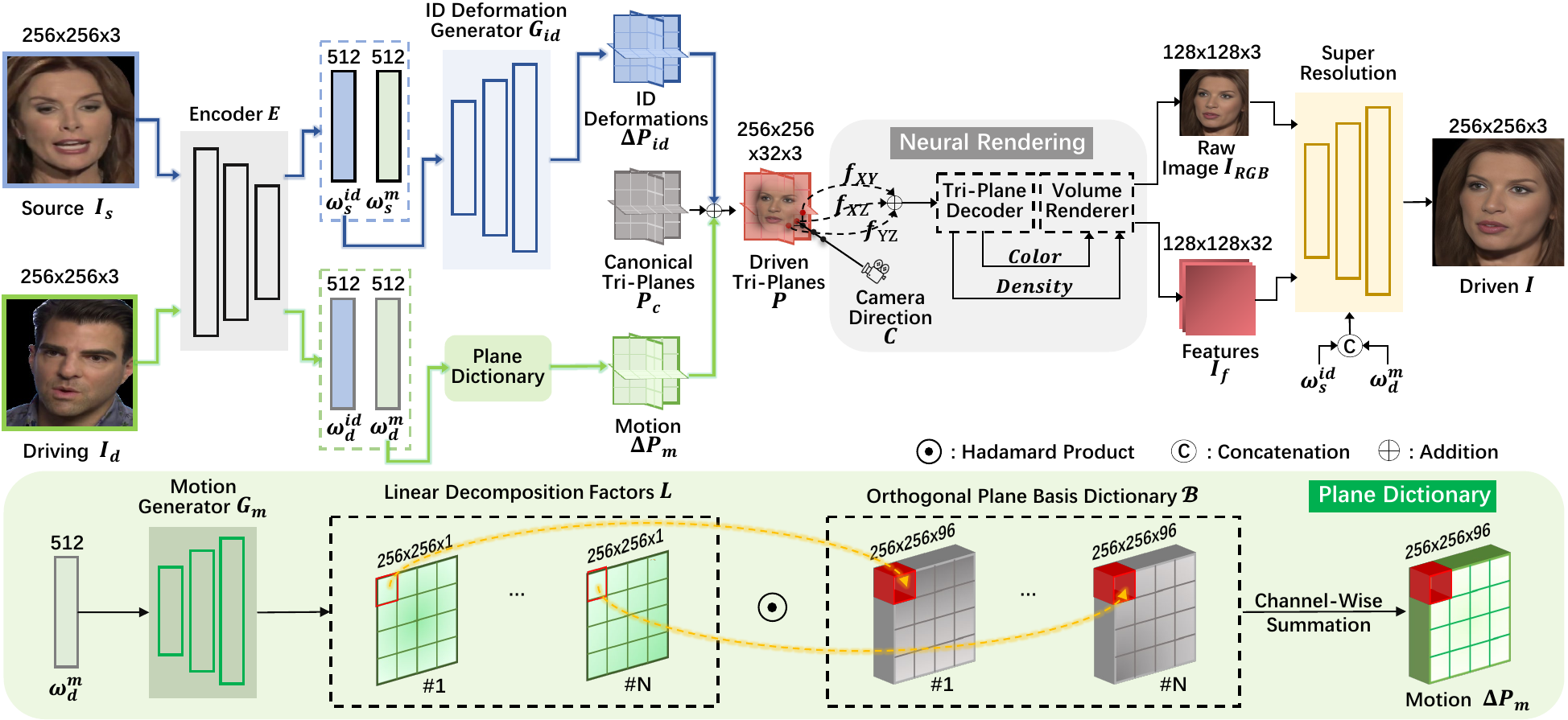}
    \caption{Method overview. We decompose face tri-planes into three components: canonical tri-planes, identity deformations, and motion. In terms of motion control, which is the topology transformation rules shared by different identities, we propose a Plane Dictionary (PlaneDict) module to transfer motion conditions between different face tri-planes for face reenactment.}
    \label{pipeline}
    
\end{figure*}

\subsection{One-Shot Face Reenactment}

Previous face reenactment methods can be divided into warping-based, mesh-based, and NeRF-based. Warping-based methods~\cite{dong2018soft,geng2018warp,liu2019liquid,ha2020marionette,drobyshev2022megaportraits,zhao2022thin,wang2022latent} warp the source features by estimated motion field to transport driving expressions and poses into the source face for 2D generation. Among them, FOMM~\cite{siarohin2019first} builds a 2D motion field from the sparse keypoints detected by an unsupervised trained detector, while DaGAN~\cite{hong2022depth} incorporates the depth estimation to supplement the missing 3D geometry information in 2D motion field. Mesh-based methods employ 3DMM uses a single image to create realistic photos in a rigged mesh format such as ROME~\cite{khakhulin2022realistic}. In terms of NeRF-based methods~\cite{guo2021ad,liu2022semantic,shen2022dfrf,li2023efficient}, using one image to build a 3D implicit representation is an ill-posed problem, because the lack of multi-view information makes the failure of learning the dense spatial information from one image. FDNeRF~\cite{zhang2022fdnerf} relaxes the constraint to the required number of images, while FNeVR~\cite{zeng2022fnevr} takes the merits of 2D warping and 3D neural rendering. As for the motion control of tri-plane representation, OTAvatar~\cite{ma2023otavatar} designs a motion encoding strategy for pre-trained EG3D~\cite{chan2022efficient} with the 3DMM prior, while HiDeNeRF~\cite{Li_2023_CVPR} proposes a multi-scare tri-plane feature extractor, as well as 3DMM-based implicit motion-conditioned deformation field, to train a generative model from scratch. However, these 3DMM-based methods still suffer from the limitations brought by 3DMM itself. Therefore, we aim to tackle the more tricky challenge that is learning the dense correspondence between different tri-planes without 3DMM prior and achieving matchable or even better results than previous methods, which can have the potential of animating arbitrary objects which lack a sophisticated 3D parametric modeling like human faces.

\section{Method}

We propose a novel framework to achieve one-shot multi-view face reenactment, which can learn the dense correspondence between different face tri-planes and realize motion transfer. In terms of motion control, our key contribution is to construct a Plane Dictionary (\textbf{PlaneDict}) module to efficiently map motion conditions to feature deformations of tri-planes, which realizes fine-grained motion transfer.

\subsection{Overview}

\noindent
\textbf{Tri-Plane Representation.} Previous studies have utilized NeRF~\cite{mildenhall2021nerf} as fundamental implicit representation. Notably, 3D generative models like StyleNeRF~\cite{gu2021stylenerf} and EG3D~\cite{chan2022efficient} combine NeRF-based representation with StyleGAN-based~\cite{karras2020analyzing} generator. These models extend identity-specific overfitting of vanilla NeRFs to a GAN space which can render 3D-consistent multi-identity face images with diverse expressions and poses. Among these NeRF variants, the tri-planes proposed by EG3D achieve a superior balance between information density and photo-realism, while also enabling the construction of a diverse latent space for manipulation. In contrast to the vanilla NeRF, which employs an MLP network to record spatial points in the space, EG3D adopts a latent vector to represent it and maps it to three spatially-orthogonal plane feature maps (i.e., tri-planes) through a StyleGAN-based generator. The tri-planes effectively provide sufficient information for rendering a spatial point and can be queried efficiently. Consequently, we adopt tri-planes as our fundamental representation.

\noindent
\textbf{Decomposition Strategy.} The computer graphics researchers~\cite{blanz1999morphable,li2017learning} typically decompose the canonical space (also known as the template mesh) and vertex deformations caused by identity and motion conditions in order to enhance the stability and interpretability of the learning process used in the mesh-based pipeline for modeling human faces. Unfortunately, the EG3D pipeline did not incorporate this decomposition structure, resulting in a lack of explicit control over the identity and motion of human faces. This limitation poses challenges for downstream applications such as facial attribute editing and face reenactment. To address this issue, previous methods~\cite{lin20223d,ma2023otavatar} have employed GAN inversion to embed real face images into the latent space of EG3D, which tend to utilize the encoder of a face parametric model to obtain explicit control over identities, expressions, and poses. However, this roadmap only distills the correspondence from the parametric models and inherits their limitations. Therefore, we decompose the canonical space, identity deformations, and motion to learn dense correspondence of tri-planes for more flexible motion transfer. 

Specifically, as shown in Fig.~\ref{pipeline}, we adopt an encoder $\bm{E}$ to extract the style codes of the input image $\bm{I_{input}}$, which can be embedded as identity code $\bm{\omega^{id}}$ and motion code $\bm{\omega^{m}}$:

\begin{equation}
\footnotesize
\setlength\abovedisplayskip{1pt}
\setlength\belowdisplayskip{1pt}
    (\bm{\omega^{id}},\bm{\omega^{m}})  =\bm {E} (\bm {I_{input}}).
\end{equation}

The dense correspondence and motion transfer are achieved by identity deformations $\Delta \bm{P_{id}}$ and motion $\Delta \bm{P_{m}}$. We feed the identity code $\bm{\omega^{id}}$ into a StyleGAN-based generator $\bm{G_{id}}$ to obtain the identity deformations $\Delta \bm{P_{id}}$. However, the motion $\Delta \bm{P_{m}}$ aims to achieve motion transfer between different face tri-planes, which means a shared deformation method should be proposed to handle this challenge. Therefore, we propose our \textbf{PlaneDict} module to obtain the motion $\Delta \bm{P_{m}}$, which will be presented in the next section. 

The tri-planes $\bm{P}$ which represents the driven face image to be rendered consist of the learnable canonical tri-planes $\bm{P_c}$, the identity deformations $\Delta \bm{P_{id}}$ from the source face image $\bm {I_{s}}$, and the motion $\Delta \bm{P_{m}}$ from the driving face image $\bm {I_{d}}$, which can be formulated as follows:

\begin{equation}
\footnotesize
\setlength\abovedisplayskip{1pt}
\setlength\belowdisplayskip{1pt}
    \bm {P}(\bm{\omega^{id}_{s}},\bm{\omega^{m}_{d}}) =\bm {P_c} + \bm {\Delta P_{id}} + \bm {\Delta P_{m}},
\end{equation}

\begin{equation}
\footnotesize
\setlength\abovedisplayskip{1pt}
\setlength\belowdisplayskip{1pt}
    \bm{\Delta P_{id}} = \bm{G_{id}}(\bm{\omega^{id}_{s}}),
\end{equation}

\begin{equation}
\footnotesize
\setlength\abovedisplayskip{1pt}
\setlength\belowdisplayskip{1pt}
    \bm{\Delta P_{m}} = \bm{PlaneDict}\;(\bm{G_{m}}(\bm{\omega^{m}_{d}});\mathcal{B}),
\end{equation}
where $\bm{G_{m}}$ and $\mathcal{B}$ are the motion deformation generator and the learnable plane bases of the PlaneDict module.

 Finally, when we query a spatial point according to its location $(x,y,z)$ and the camera direction $C$, we sample features ($\bm{f_{XY}}$, $\bm{f_{XZ}}$, and $\bm{f_{YZ}}$) from the driven tri-planes $\bm{P}$, aggregate by summation, and process the aggregated features with a lightweight tri-plane decoder. This decoder outputs a scalar density and a 32-channel feature, and both of them are then processed by a volume renderer to project the 3D feature volume into a 2D feature image. For training efficiency, we render 32-channel feature maps $\bm{I_f}$ at a resolution of $128^2$, with $96$ total depth samples per ray. And we adopt the Super Resolution module to increase the final image size to $256^2$, which utilizes two blocks of StyleGAN-modulated convolution layers that upsample and refine the 32-channel feature maps $\bm{I_f}$ into the final RGB image $\bm{I}$.

\subsection{Plane Dictionary (PlaneDict)}


\noindent
\textbf{Preliminary.} In the 3D Morphable Model (3DMM)~\cite{blanz1999morphable}, every face is represented by a shared topology which consists of vertexes with the aligned index. The hundreds of these 3D faces are high dimensional data and then reduced through Principal Component Analysis (PCA) to several orthogonal vector bases. These bases are further divided into identity-related and expression-related bases. When we fit the mesh model to a target face, we only need to linearly add these orthogonal bases to obtain the identity and expression deformations of every vertex relative to the template mesh, and finally get the 3D representation of the target face. It is worth noting that the expression representation of different faces can be obtained through linear addition of expression bases, which means that through the above modeling method, we can achieve dense correspondence between different face representations. 

\noindent
\textbf{Motivation.} The modeling approach of graphics inspires us to establish dense correspondence and realize motion transfer between different face tri-planes. However, if we have achieved such dense correspondence, how can we transform these motion conditions into deformations of each implicit spatial point? The previous methods~\cite{lin20223d,Li_2023_CVPR,ma2023otavatar} either skipped this issue and directly learned the mapping relationship between 3DMM and their latent space of generative models, and then used dense correspondence of 3DMM to control their own generative model; or conducted the learning of dense correspondence and motion transfer in the latent vector space corresponding to each implicit representation. The former is limited by 3DMM and cannot handle non-facial areas such as hair and eyes; The latter, which ignores the inherent characteristics of implicit representation, cannot perform more fine-grained expression control. Therefore, we propose a Plane Dictionary (\textbf{PlaneDict}) module, which can obtain the plane feature deformations driven by motion conditions by linearly adding a set of orthogonal plane bases. 

Specifically, as shown in Fig.~\ref{pipeline} and Eqn.~(4), the driving motion code $\bm{\omega^{m}_{d}}$ is first fed into the motion generator $\bm{G_{m}}$ to obtain the linear decomposition factors $\bm L$, which consists of $N$ feature maps. These decomposition factors are then multiplied by the orthogonal plane bases in the plane dictionary $\mathcal{B}$ through the Hadamard product. The plane bases in $\mathcal{B}$ are channel-wise orthogonal, i.e., $N$ vectors that have the same channel index in these orthogonal plane feature maps are reduced by QR decomposition to maintain the orthogonality, and they are learnable in the training stage. Finally, the Hadamard product of $\bm L$ and $\mathcal{B}$ is channel-wisely summed to output the motion $\Delta \bm{P_{m}}$. Note that our face motion conditions include facial expressions and head poses.

\subsection{Optimization}

The goal of proposing our framework is to learn the dense correspondence. We have the assumption that different identities have different topological structures which is suitable for modeling by a StyleGAN-based generator, and the rules of topological transformations are shared among different identities which can be modeled by our PlaneDict module. So we only adopt a self-supervision manner to train our framework. If this self-supervision successfully disentangles the deformations brought by identity and motion conditions, it precisely indicates that our aforementioned assumptions are valid, and our framework can exactly learn dense correspondence without a 3D parametric model prior.

The cross-identity driving is our ultimate goal, and we should fully utilize the paired data and learn the motion deformations with ground truth. Therefore, we adopt disentanglement loss $\mathcal{L}_{dis}$ to model the different deformations brought identity and motion conditions, and reconstruction loss $\mathcal{L}_{recons}$ to improve the image quality and rendering details as follows:

\begin{equation}
\footnotesize
\setlength\abovedisplayskip{1pt}
\setlength\belowdisplayskip{1pt}
    \mathcal{L} =\mathcal{L}_{dis} + \lambda_{1}\mathcal{L}_{recons}.
\end{equation}

\noindent
\textbf{Disentanglement Loss.} We denote the input source image as $\bm{I_{s}}$, the input driving image as $\bm{I_{d}}$ and the output driven image as $\bm{I}$. We adopt the encoder $\bm{E}$ to extract $\bm{\omega^{id}_{s}}$ \& $\bm{\omega^{id}_{d}}$ and $\bm{\omega^{m}_{s}}$ \& $\bm{\omega^{m}_{d}}$ from $\bm{I_{s}}$ and $\bm{I_{d}}$ respectively. Moreover, we use $\bm{E}$ to extract the $\bm{\omega^{id}}$ \& $\bm{\omega^{m}}$ from $\bm{I}$. Our optimization goal is to minimize the distance between $\bm{\omega^{id}_{s}}$ and $\bm{\omega^{id}}$, and the distance between $\bm{\omega^{m}_{d}}$ and $\bm{\omega^{m}}$. Therefore, we propose our disentanglement loss $\mathcal{L}_{dis}$:

\begin{equation}
\footnotesize
\setlength\abovedisplayskip{1pt}
\setlength\belowdisplayskip{1pt}
    \mathcal{L}_{dis} = \Vert \bm{\omega^{id}_{s}}-\bm{\omega^{id}} \Vert_{2} + \lambda_{2}\Vert \bm{\omega^{m}_{d}}-\bm{\omega^{m}} \Vert_{2}.
\end{equation}

\noindent
\textbf{Reconstruction Loss.} When we conduct the experiments, we found that the regions of eyes and mouths take longer training time to learn the distribution, which are high-frequency details with small areas and big variations. Therefore, as for the reconstruction loss $\mathcal{L}_{recons}$, we include $\mathcal{L}_{1}$ loss and mask loss, as follows:

\begin{equation}
\footnotesize
\setlength\abovedisplayskip{1pt}
\setlength\belowdisplayskip{1pt}
    \mathcal{L}_{recons} = \Vert \bm{I_{d}}-\bm{I} \Vert_{1} + \lambda_{3}\Vert \mathcal{M}(\bm{I_{d}})-\mathcal{M}(\bm{I}) \Vert_{1},
\end{equation}

\noindent
where $\mathcal{M}$ is a mask that indicates the eye and mouth part of the face image.

\noindent
\textbf{Training Strategy.} We use two types of paired data and use different losses to optimize the network parameters. One type is to use the source image and target image from the same identity, and we adopt $\mathcal{L}$ for training. The other is to use the source image and target image from different identities, and we only use $\mathcal{L}_{dis}$ for training.

\section{Experiments}
\subsection{Experimental Settings}

\begin{figure}[htb]
    \centering
    \includegraphics[scale=0.57]{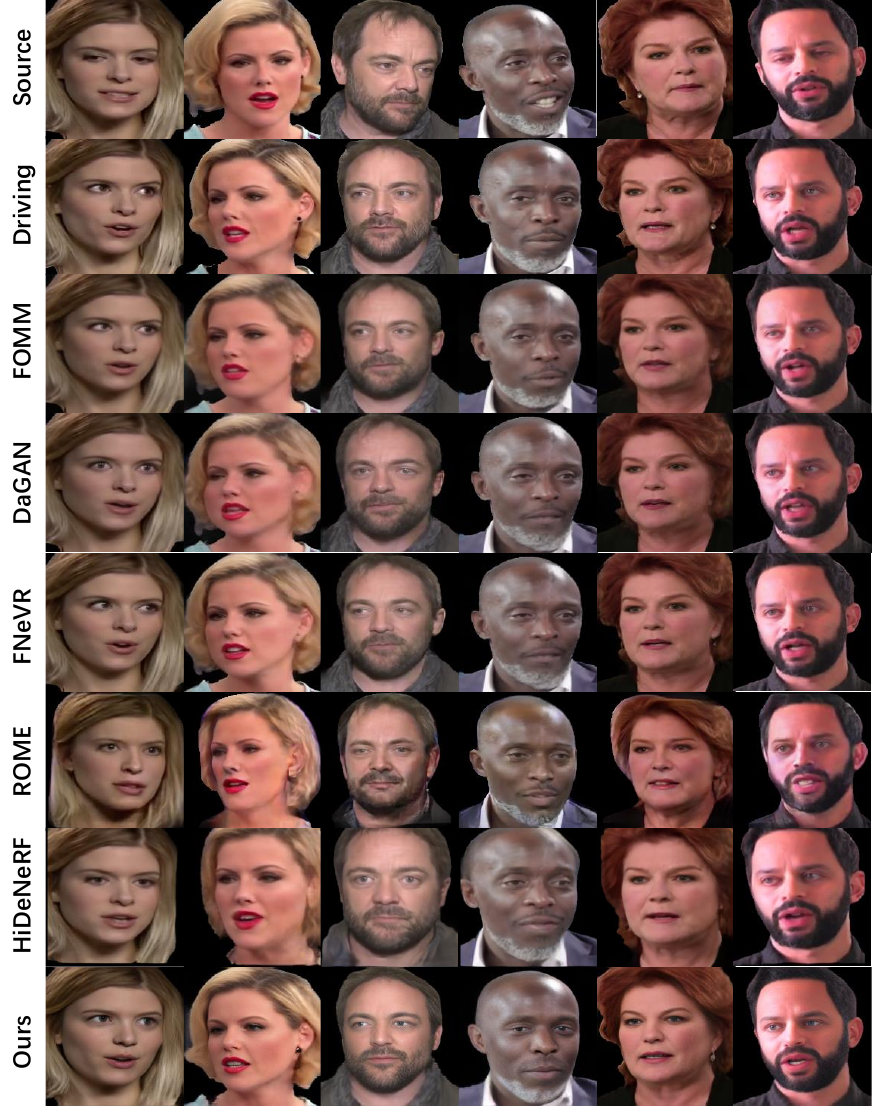}
    \caption{Qualitative results of self-reenactment.}
    \label{self-reenactment}
    
\end{figure}

\noindent
\textbf{Implementation Details.} The encoder $\bm{E}$ is a ResNet-10~\cite{he2016deep} network. The generators $\bm{G_{id}}$ and $\bm{G_{m}}$ are two StyleGAN-based generators~\cite{karras2020analyzing}. The Super Resolution module utilizes two blocks of StyleGAN-modulated layers. To further improve the image resolution and quality, we adopt the pre-trained dual discriminator from EG3D~\cite{chan2022efficient} and sample data from FFHQ~\cite{karras2019style} as real images to fine-tune our networks in a GAN manner. The $\lambda_{1}$, $\lambda_{2}$, and $\lambda_{3}$ are set as 1.0, 1.5, and 10. The iteration ratio of self-reenactment and cross-identity reenactment is better set at 2:1. Using Adam optimizer (set learning rate as 0.0001), the training takes about 4 days on 8 Tesla V100 GPUs while the fine-tuning takes 1 day.

\noindent
\textbf{Baselines.} We select five state-of-the-art methods from different perspectives, including 2D-warping-based FOMM~\cite{siarohin2019first} \& DaGAN~\cite{hong2022depth}, mesh-based ROME~\cite{khakhulin2022realistic}, NeRF-based FNeVR~\cite{zeng2022fnevr}, and tri-plane-based HiDeNeRF~\cite{Li_2023_CVPR}. For fair comparisons, these methods are trained with VoxCeleb dataset~\cite{nagrani2017voxceleb,chung2018voxceleb2}.

\noindent
\textbf{Datasets.} We conduct experiments over three commonly used datasets: VoxCeleb1~\cite{nagrani2017voxceleb}, VoxCeleb2~\cite{chung2018voxceleb2}, and TalkingHead-1KH~\cite{wang2021one}. We follow the FOMM to pre-process these videos, in which each frame is aligned and cropped into $256^2$ resolution. We follow the EG3D~\cite{chan2022efficient} to extract camera extrinsics, which is based on an off-the-shelf pose estimator~\cite{deng2019accurate}. Furthermore, we use face-parsing.Pytorch~\cite{faceparsing} to provide region masks of face, hair, and torso, and set the background region as black, which can reduce the impact of complex backgrounds. The selected videos for the test are not overlapped with the training videos.

\begin{table}[htb]
\setlength\tabcolsep{3pt}
\footnotesize
        \centering
        \begin{tabular}{cccc}
        \toprule[1pt]
        ~&SSIM$\uparrow$&PSNR$\uparrow$&LPIPS$\downarrow$\\
        \midrule[0.05pt]
        FOMM&0.690&19.2&0.112\\
        DaGAN&0.807&23.2&0.088\\
        FNeVR&0.901&21.1&0.092\\
        ROME&0.833&21.6&0.085\\
        HiDeNeRF&0.862&21.9&0.084\\
        Ours&\textbf{0.870}&\textbf{22.1}&\textbf{0.079}\\
        \bottomrule[1pt]
        \end{tabular}
        \caption{Visual quality evaluation of self-reenactment.}
        \label{quality_self}
        
    \end{table}
    
    \begin{table}[htb]
\setlength\tabcolsep{3pt}
\footnotesize
        \centering
        \begin{tabular}{cccccc}
        \toprule[1pt]
        ~&CSIM$\uparrow$&AUCON$\uparrow$&PRMSE$\downarrow$&AVD$\downarrow$&ET$\downarrow$\\
        \midrule[0.05pt]
        FOMM&0.837&0.872&2.88&0.021&1.98\\
        DaGAN&0.875&0.921&1.79&0.016&4.08\\
        FNeVR&0.880&0.929&2.22&0.016&2.01\\
        ROME&0.906&0.918&1.68&0.013&5.28\\
        HiDeNeRF&0.931&0.956&1.66&0.010&5.44\\
        Ours&\textbf{0.946}&\textbf{0.961}&\textbf{1.60}&\textbf{0.009}&\textbf{1.72}\\
        \bottomrule[1pt]
        \end{tabular}
        \caption{Identity fidelity and motion accuracy evaluation of self-reenactment.}
        \label{fidelity_self}
        
    \end{table}

\noindent
\textbf{Metrics.} We evaluate different methods from three perspectives: (1) Visual quality: We adopt SSIM~\cite{wang2004image}, PSNR, LPIPS~\cite{zhang2018unreasonable}, and FID~\cite{heusel2017gans} as quality metrics. (2) Identity fidelity and motion accuracy: Following the previous works~\cite{ha2020marionette,hong2022depth}, we adopt CSIM, PRMSE, and AUCON to evaluate the identity preservation of the source image, the accuracy of head poses, and the precision of expression. (3) Multi-view consistency: We adopt the AVD proposed by HiDeNeRF~\cite{Li_2023_CVPR} to evaluate multi-view identity preservation. Furthermore, we propose the ET (Eye Tracking) metric to evaluate fine-grained motion control of gaze, which calculate the error of eye locations~\cite{gazetracking} between the source image and driving image (they are all aligned face images).

\subsection{Self-Reenactment}

The self-reenactment experiments are using the source and driving images of the same identity, which have the ground truth of the synthesized results for comparisons. As shown in Fig.~\ref{self-reenactment}, we show the qualitative results of different methods. Because we use the motion features extracted from the driving images, instead of 3DMM parameters, we can not only realize more fine-grained motion control than 3DMM-based methods but also handle special regions like hair and eyes. We list the quantitative results in Tab.~\ref{quality_self} and Tab.~\ref{fidelity_self}, and we achieve matched or better scores than other state-of-the-art methods which are based on the correspondence from the 3DMM prior. These results show that, instead of using 3DMM parameter control at the cost of missing details, we can directly establish dense correspondence between different tri-plane representations, which overcomes the optimization burden of aligning 3DMM space and the latent space of NeRF-based generative models.

\begin{figure}[htb]
    \centering
    \includegraphics[scale=0.58]{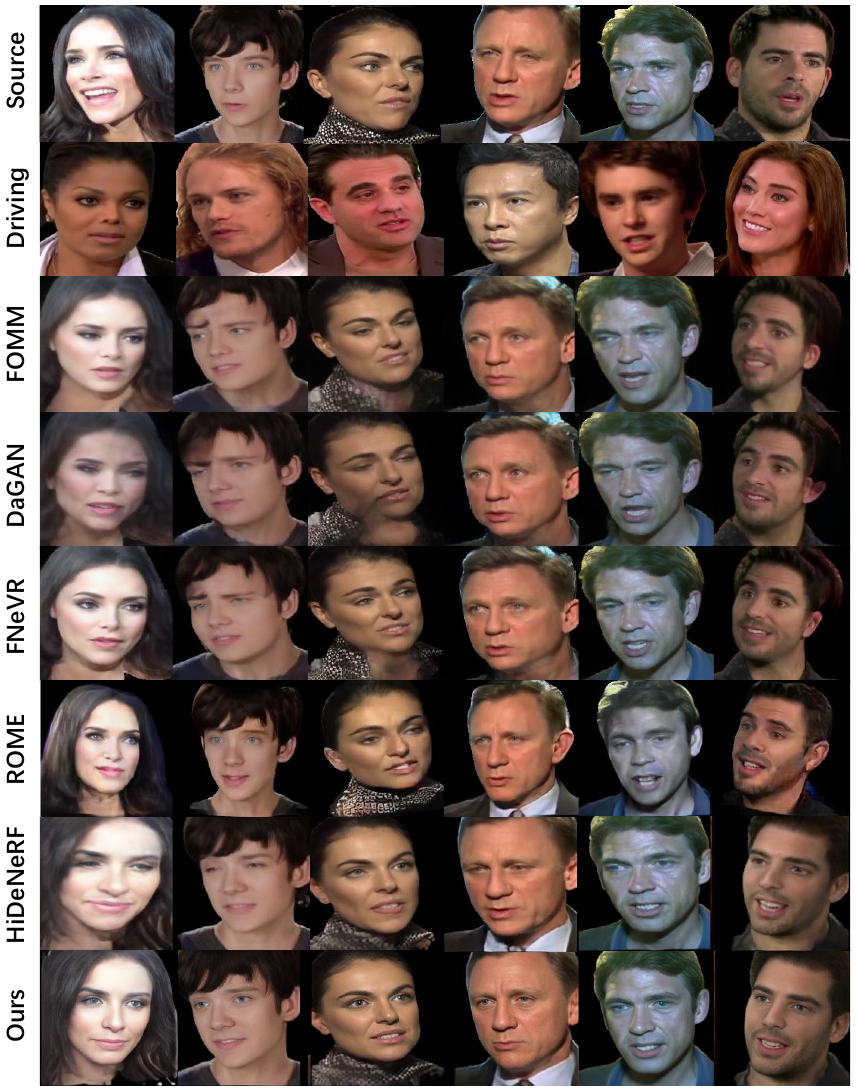}
    \caption{Qualitative results of cross-identity reenactment.}
    \label{cross-identity}
    
\end{figure}
    
\subsection{Cross-Identity Reenactment}

The cross-identity reenactment is using the source and driving images of different identities, which is a more difficult challenge for source identity preservation and fine-grained motion transfer between different face tri-planes. We first qualitatively compare different state-of-the-art methods and

\begin{figure}[H]
    \centering
    \includegraphics[scale=0.355]{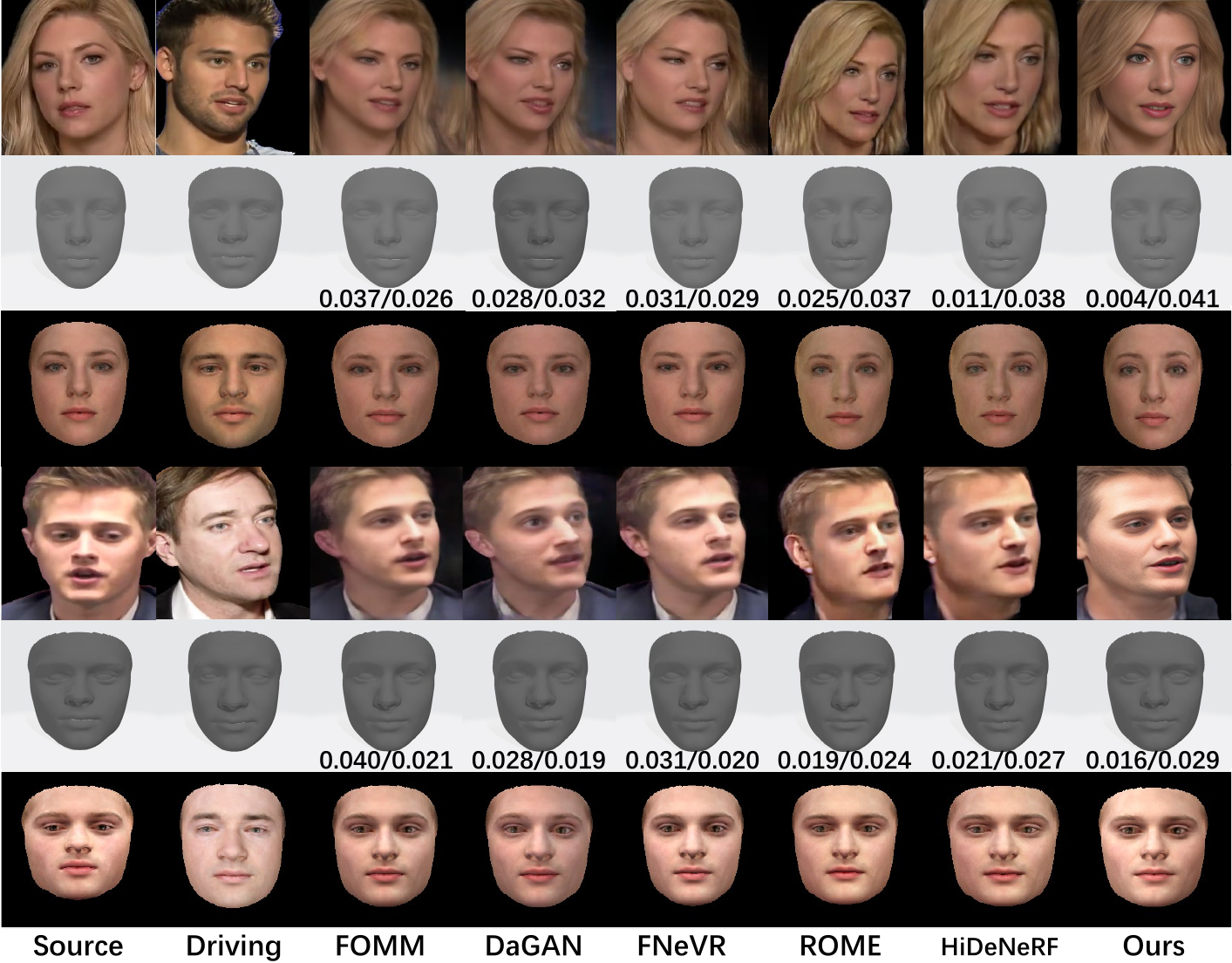}
    \caption{Mesh evaluation for identity preservation of cross-identity reenactment (vertex distance from source identity$\downarrow$/ vertex distance from driving identity$\uparrow$).}
    \label{mesh_visualization}
    \end{figure}

    \begin{table}[htb]
        \setlength\tabcolsep{3pt}
        \footnotesize
        \centering
        \begin{tabular}{ccccccc}
        \toprule[1pt]
        \multicolumn{1}{c}{\multirow{2}*{}}&\multicolumn{5}{c}{VoxCeleb1}\\
        ~&CSIM$\uparrow$&AUCON$\uparrow$&PRMSE$\downarrow$&FID$\downarrow$&AVD$\downarrow$&ET$\downarrow$\\
        \midrule[0.05pt]
        FOMM&0.748&0.752&3.66&86&0.044&6.08\\
        DaGAN&0.790&0.880&3.06&87&0.036&6.16\\
        FNeVR&0.812&0.884&3.32&82&0.041&6.10\\
        ROME&0.833&0.871&2.64&76&0.016&7.08\\
        HiDeNeRF&0.876&0.917&2.62&57&0.012&7.02\\
        Ours&\textbf{0.911}&\textbf{0.928}&\textbf{2.50}&\textbf{49}&\textbf{0.011}&\textbf{5.18}\\
        
        \midrule[0.05pt]
        \multicolumn{1}{c}{\multirow{2}*{}}&\multicolumn{5}{c}{VoxCeleb2}\\
        ~&CSIM$\uparrow$&AUCON$\uparrow$&PRMSE$\downarrow$&FID$\downarrow$&AVD$\downarrow$&ET$\downarrow$\\
        \midrule[0.05pt]
        FOMM&0.680&0.707&4.16&85&0.047&6.23\\
        DaGAN&0.693&0.815&3.93&86&0.040&6.62\\
        FNeVR&0.699&0.829&3.90&84&0.047&5.99\\
        ROME&0.710&0.821&3.08&76&0.019&7.29\\
        HiDeNeRF&0.787&0.889&2.91&61&0.014&7.30\\
        Ours&\textbf{0.790}&\textbf{0.894}&\textbf{2.83}&\textbf{58}&\textbf{0.012}&\textbf{5.33}\\

        \midrule[0.05pt]
        \multicolumn{1}{c}{\multirow{2}*{}}&\multicolumn{5}{c}{TalkingHead-1KH}\\
        ~&CSIM$\uparrow$&AUCON$\uparrow$&PRMSE$\downarrow$&FID$\downarrow$&AVD$\downarrow$&ET$\downarrow$\\
        \midrule[0.05pt]
        FOMM&0.723&0.741&3.71&76&0.039&6.17\\
        DaGAN&0.766&0.872&2.98&73&0.035&6.59\\
        FNeVR&0.775&0.879&3.39&73&0.037&6.03\\
        ROME&0.781&0.864&2.66&68&0.017&6.97\\
        HiDeNeRF&0.828&0.901&2.60&52&0.011&7.09\\
        Ours&\textbf{0.831}&\textbf{0.912}&\textbf{2.55}&\textbf{49}&\textbf{0.010}&\textbf{5.42}\\
        
        \bottomrule[1pt]
        \end{tabular}
        \caption{Cross-identity reenactment evaluation.}
        \label{cross}
        
    \end{table}

     \noindent
show their synthesized results in Fig.~\ref{cross-identity}. Although the fusion of identity and motion information is a hard problem, our framework with the PlaneDict module is able to generate cross-identity reenactment results with better image quality and identity fidelity without any artifacts. To further compare the identity preservation ability of our framework, as shown in Fig.~\ref{mesh_visualization}, we evaluate and visualize the identity fidelity (i.e., the shape preservation of the source identity) using 3D face reconstruction models~\cite{deng2019accurate}. The quantitative experiments are performed with VoxCeleb1~\cite{nagrani2017voxceleb}, VoxCeleb2~\cite{chung2018voxceleb2}, and TalkingHead-1KH~\cite{wang2021one}, which are shown in Tab.~\ref{cross} respectively.

\subsection{Multi-View Synthesis}
\begin{figure}[htb]
    \centering
    \includegraphics[scale=0.34]{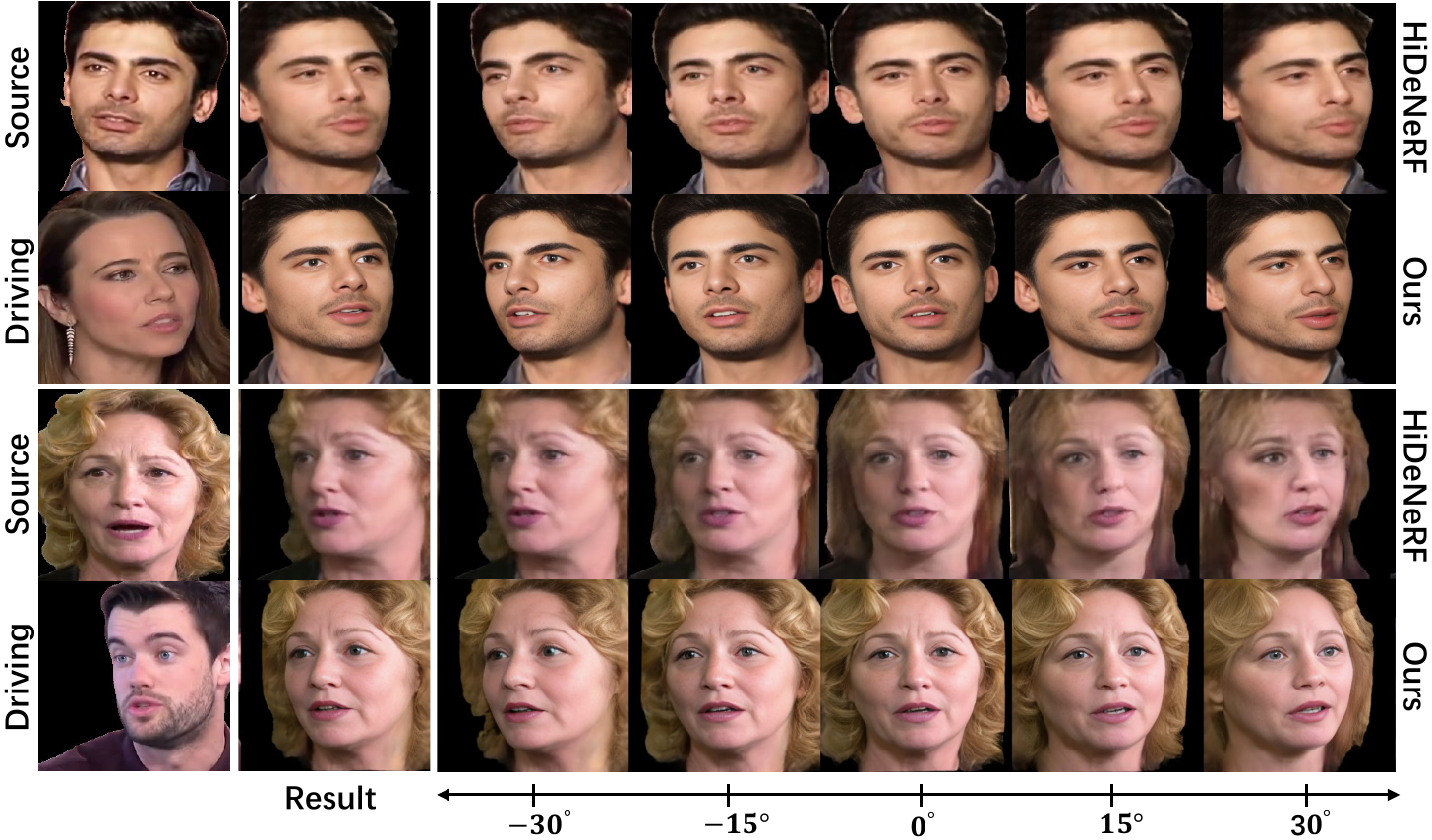}
    \caption{Qualitative results of multi-view synthesis.}
    \label{multi}
    
    \end{figure}

The one-shot multi-view cross-identity reenactment is the most challenging task. It requires not only using one face image from the source identity to construct a 3D face representation for multi-view rendering but also this representation can be controlled by motion conditions for novel expression and pose reenactment. We adopt the state-of-the-art HiDeNeRF~\cite{Li_2023_CVPR} as the baseline method for comparison. As shown in Fig.~\ref{multi}, we render the driven results from different view directions. Our method achieves better image quality than HiDeNeRF and does not show artifacts or 3D inconsistency in some angles. As shown in Tab.~\ref{multiview}, we further evaluate ours and HiDeNeRF in quantitative experiments. Our framework with the PlaneDict module can replace the 3DMM to model the dense correspondence between different tri-plane representations. In this way, instead of aligning the implicit spaces of the two generative models, we learn the dense correspondence without any loss of the 3D consistency of the target NeRF-based model.

\begin{table}[htb]
    \setlength\tabcolsep{3pt}
    \footnotesize
            \centering
            \begin{tabular}{ccccc}
            \toprule[1pt]
            ~&CSIM$\uparrow$&AUCON$\uparrow$&PRMSE$\downarrow$&AVD$\downarrow$\\
            \midrule[0.05pt]
            HiDeNeRF&0.829&0.864&3.78&0.014\\
            Ours&\textbf{0.840}&\textbf{0.881}&\textbf{3.53}&\textbf{0.008}\\
            \bottomrule[1pt]
            \end{tabular}
            \caption{Quantitative evaluation of multi-view synthesis.}
            \label{multiview}
            
        \end{table}

        \begin{table}[H]
            \setlength\tabcolsep{3pt}
            \footnotesize
            \centering
            \begin{tabular}{cccccc}
            \toprule[1pt]
            ~&CSIM$\uparrow$&AUCON$\uparrow$&PRMSE$\downarrow$&AVD$\downarrow$&ET$\downarrow$\\
            \midrule[0.05pt]
            w/o PlaneDict&0.763&0.809&3.10&0.035&7.58\\
            w PlaneDict (5)&0.679&0.718&3.93&0.058&9.92\\
            w PlaneDict (10)&0.802&0.824&3.18&0.038&7.36\\
            w PlaneDict (15)&0.899&0.871&2.92&0.019&6.69\\
            w PlaneDict (20)&\textbf{0.911}&\textbf{0.928}&\textbf{2.50}&\textbf{0.011}&\textbf{5.18}\\
            \bottomrule[1pt]
            \end{tabular}
            \caption{Ablation study.}
            \label{ablation}
            
        \end{table}

    \subsection{Ablation Study}
As shown in Tab.~\ref{ablation}, in the ablation study of whether to use the PlaneDict module, we adopt the same structure as identity deformations for obtaining motion and they are trained with the same time and dataset. Limited by the hardware, the max number of plane bases is 23. However, since 20, there has been almost no improvement. Therefore, we adopt 20 as the number of plane bases in our PlaneDict module to balance the quality and optimization difficulty.

\section{Conclusions}

In this paper, we propose a novel framework to learn the dense correspondence between different face tri-planes without a 3D parametric model prior. With the PlaneDict module, our framework can achieve fine-grained motion driving of face tri-planes without any 3D inconsistency. Extensive experiments demonstrate our better image quality, fine-grained motion control, and identity fidelity of one-shot multi-view face reenactment than previous methods. 

\noindent
\textbf{Limitations and Ethical Concerns.} Due to the inherent biases in the datasets, we are not able to handle extreme poses and expressions. We strongly oppose any misuse of our technology but we believe it has the potential to achieve multi-view animation of diverse objects without relying on sophisticated 3D parametric models like human faces.

\section*{Acknowledgments}

This work is supported by the National Key Research and Development Program of China under Grant No. 2021YFC3320103, the National Natural Science Foundation of China (NSFC) under Grants 62372452, U19B2038.

\bibliography{aaai24}

\end{document}